\documentclass{article}
\usepackage[numbers]{natbib}
\usepackage{spconf,amsmath,graphicx,hyperref}
\usepackage[T1]{fontenc}
\usepackage{verbatim}
\usepackage{amssymb}
\usepackage{booktabs}
\usepackage{multirow}
\usepackage{xcolor}
\usepackage{tikz}
\usepackage{pgfplots}
\pgfplotsset{compat=1.17}
\usepackage{array}
\usepackage{tabularx}
\usepackage{url}
\usepackage{microtype}
\usepackage{enumitem}
\usepackage{subcaption}

\title{MINT: Multimodal Imaging-to-Speech Knowledge Transfer for \\Early Alzheimer's Screening}

\name{Vrushank Ahire$^{1,\dagger}$ \quad
Yogesh Kumar$^{1,\dagger}$ \quad
Anouck Girard$^{2}$ \quad
M. A. Ganaie$^{1}$%
\thanks{$^\dagger$Equal contribution}}

\address{%
\normalsize
$^{1}$Department of Computer Science and Engineering,
Indian Institute of Technology Ropar, India\\
\normalsize$^{2}$Embry-Riddle Aeronautical University, Florida, USA\\
\normalsize{2022csb1002@iitrpr.ac.in, yogesh.23csz0014@iitrpr.ac.in, girarda3@erau.edu, mudasir@iitrpr.ac.in
}}

\begin{document}

\maketitle

\begin{abstract}
Alzheimer's disease is a progressive neurodegenerative disorder in which mild cognitive impairment (MCI) precedes dementia. Structural MRI provides biomarkers but requires costly infrastructure, limiting population-scale deployment. Speech offers a non-invasive alternative, yet speech-only classifiers are developed independently of neuroimaging and lack biological grounding for CN-versus-MCI classification. We propose MINT (Multimodal Imaging-to-Speech Knowledge Transfer), a three-stage framework that transfers MRI-derived biomarker structure to speech during training. An MRI teacher defines a compact embedding space for CN-versus-MCI classification, while a residual projection head aligns speech representations to this space using a combined geometric loss. The frozen MRI classifier enables imaging-free inference. On ADNI-4, aligned speech achieves performance comparable to speech baselines, while multimodal fusion improves over MRI alone. Ablations identify dropout regularization and self-supervised pretraining as important design choices. To our knowledge, MINT is the first demonstration of MRI-to-speech knowledge transfer for early Alzheimer's screening without imaging at inference.
\end{abstract}

\section{Introduction}
\label{sec:intro}

Alzheimer’s disease (AD) and related dementias affect over 55 million people worldwide, and mild cognitive impairment (MCI) represents a critical stage before dementia~\cite{world2021global}. Early detection can improve clinical management, yet population-scale screening remains challenging. Structural MRI provides established biomarkers of neurodegeneration, but its cost, specialised expertise, and infrastructure requirements limit deployment in primary-care and low-resource settings~\cite{jack2018nia}.

Deep learning on structural MRI has achieved reliable MCI and AD classification using hippocampal and whole-brain representations~\cite{dong2025convolutional,yuan2025magnetic}. Self-supervised and multimodal learning further improve representation quality under limited clinical labels~\cite{thrasher2024te,li2024joint}. However, these approaches still require neuroimaging at inference. 

Speech provides a scalable, non-invasive alternative. Narrative and verbal-fluency tasks capture lexical, prosodic, and articulatory changes associated with early neurodegeneration~\cite{fraser2015linguistic,chandler2023explainable}. Acoustic features combined with self-supervised speech encoders have shown competitive performance on dementia benchmarks~\cite{zhu2021wavbert,li2024useful}. However, most speech models are trained independently of neuroimaging biomarkers, limiting biological grounding. Multimodal fusion can exploit complementary information~\cite{lin2024multimodal}, but generally requires both modalities at test time.

Knowledge distillation and cross-modal representation learning allow a student modality to inherit information from a teacher representation while operating independently at inference~\cite{hinton2015distilling}. Such strategies have been explored across imaging, audio--visual, and vision--language settings~\cite{kunanbayev2024training,radford2021learning}; however, transferring a neuroimaging-derived biomarker space to speech for CN-versus-MCI detection remains unexplored. We address this gap with MINT, which uses an MRI teacher to define a biomarker space predictive of MCI and trains a speech projection head to align with it. This enables speech-based inference without MRI while retaining the option of multimodal fusion. 

\begin{figure*}[h!]
    \centering
    \includegraphics[width=\textwidth]{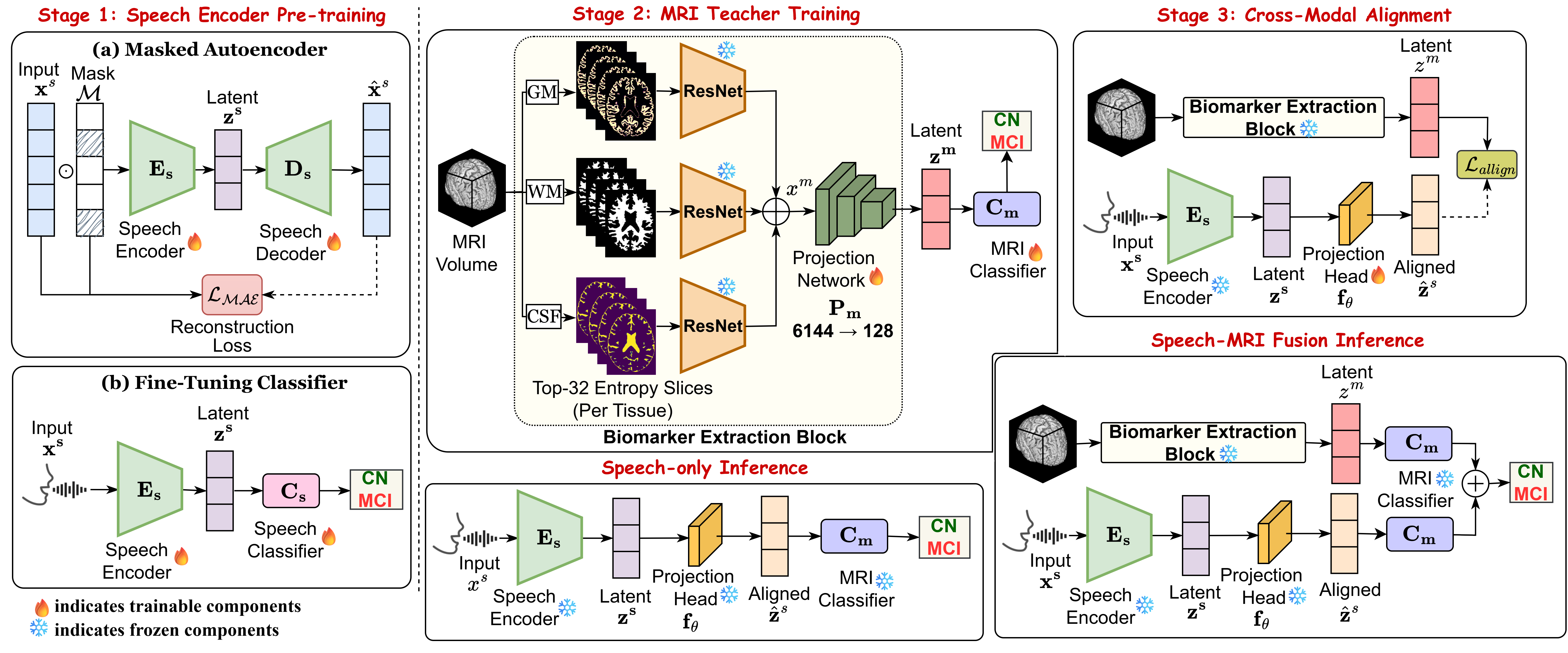}
    \caption{Overview of the three-stage MINT framework. In Stage 1, a speech encoder is pretrained using masked autoencoding and then fine-tuned for CN–MCI classification. In Stage 2, an MRI teacher is trained via tissue-stratified feature extraction to learn a 128-dimensional biomarker embedding space. In Stage 3, a projection head aligns speech embeddings to the frozen MRI space, enabling speech-only inference as well as multimodal fusion.}
    \label{fig:pipeline}
\end{figure*}

Our contributions are a three-stage teacher--student framework combining self-supervised speech pretraining, MRI-based biomarker learning, and cross-modal speech projection; an alignment strategy that transfers imaging-derived decision structure to speech; systematic ablations of projection-head regularization and self-supervised pretraining; and evaluation showing that aligned speech achieves competitive CN-versus-MCI classification while fusion improves over MRI alone. Figure~\ref{fig:pipeline} illustrates the three-stage pipeline.

\section{Methodology}
\label{sec:method}

\subsection{Problem Setup and Notation}

Let $\mathbf{x}^s\in\mathbb{R}^{209}$ and $\mathbf{x}^m\in\mathbb{R}^{6144}$ denote speech and MRI features, respectively. Using 266 paired subjects, we learn a speech encoder and projection head for CN-MCI classification without MRI at inference. Let $\mathbf{z}^s,\mathbf{z}^m\in\mathbb{R}^{128}$ denote fine-tuned speech latents and MRI biomarker embeddings, respectively, and $\hat{\mathbf{z}}^s=f_\theta(\mathbf{z}^s)$ the aligned speech representation. The shared 128-dimensional bottleneck facilitates alignment and limits overfitting.
\subsection{Stage 1: Speech Encoder Pretraining and Fine-tuning}

\noindent\textbf{Self-supervised pretraining.}
Given that the unified paired cohort contains 266 subjects, with 186 subjects in the training partition, we first pretrain the speech encoder $E_s$ on a pool of 14{,}235 unlabeled acoustic feature samples using a masked autoencoder (MAE)~\cite{yoon2020vime} objective inspired by
BERT~\cite{devlin2019bert}.
Pretraining on unlabeled data is particularly important here because the
encoder must learn a rich, general-purpose acoustic representation before
being exposed to the small labeled set; without this initialization, supervised fine-tuning on the small labeled set risks fitting noise rather than clinically meaningful patterns.

A random subset of the 209 input features is masked; the encoder-decoder ${E_s, D_s}$ reconstructs the original input $\mathbf{x}$ as $\hat{\mathbf{x}}$ via:
\begin{equation}
  \mathcal{L}_\text{MAE} =
  \frac{\sum_{i \in \mathcal{M}} (x_i - \hat{x}_i)^2}{|\mathcal{M}|}
  \;+\; \lambda_c \cdot \bigl(1 - \cos(\mathbf{x}, \hat{\mathbf{x}})\bigr),
\label{eq:mae}
\end{equation}
where $\mathcal{M}$ denotes the set of masked indices and $\lambda_c$ is
tuned by Bayesian HPO (Optuna~\cite{akiba2019optuna}) together with
architecture hyperparameters. The cosine term in Eq.~\eqref{eq:mae}
preserves directional structure in the latent space, which improves
stability during later cross-modal alignment. Intuitively, two embeddings
that point in the same direction in $\mathbb{R}^{128}$ will respond
similarly to a linear classifier; encouraging this property during
pretraining means the encoder already organizes features in a way that
is compatible with the downstream alignment objective.

\noindent\textbf{Supervised fine-tuning.}
The pretrained encoder $E_s$ is fine-tuned for CN-vs-MCI classification
by attaching a single linear head $C_s: \mathbb{R}^{128}\!\to\!\mathbb{R}^{2}$
and optimizing with soft cross-entropy. Given the class imbalance
($\approx$2:1) and small dataset, we apply Mixup
augmentation~\cite{zhang2017mixup} with $\alpha\!=\!0.3$ and label smoothing
$\varepsilon\!=\!0.10$. Mixup creates convex combinations of training
examples and their labels, which has been shown to reduce overconfident
predictions and improve calibration under class imbalance~\cite{zhang2017mixup};
label smoothing provides a complementary regularization effect by preventing
the model from assigning zero probability to non-target classes.
Discriminative learning rates are used: $10^{-4}$ for the freshly
initialized head and $10^{-5}$ for the pretrained encoder. Using a lower
learning rate for the encoder preserves the general acoustic structure
learned during pretraining while allowing the head to adapt quickly to
the classification boundary. To address the class imbalance, inverse-frequency weighting is applied with $w_c = N/(2N_c)$.
\subsection{Stage 2: MRI Feature Extraction and Teacher Training}
\textbf{Feature extraction.}
Raw T1-weighted MRI volumes are preprocessed with N4 bias-field correction, skull stripping, and registration to MNI space using ANTs SyN~\cite{avants2011reproducible}. Tissue segmentation (CSF, gray matter, white
matter) is performed using the Atropos algorithm~\cite{avants2011open}.
For each tissue type, the 32 axial slices with the highest tissue-probability–weighted entropy are selected. An ImageNet-pretrained ResNet-50~\cite{he2016deep} extracts 2048-d feature vector per slice, which is averaged across slices~\cite{tanveer2021classification}.
Concatenating the three tissue vectors yields $\mathbf{x}^m \!\in\!
\mathbb{R}^{6144}$ for each subject. Tissue stratification is motivated by compartment-specific neurodegenerative patterns: gray matter atrophy is an early marker of MCI, whereas white matter and CSF alterations provide supporting information regarding structural decline.~\cite{jack2018nia}. Entropy-based slice selection prioritizes regions with high tissue contrast and excludes near-uniform background areas.

\noindent\textbf{MRI teacher.}
A deep MLP with monotonically decreasing hidden widths $6144 \!\to\! h_1 \!\to\! \cdots \!\to\! h_k \!\to\! 128$ is trained to classify CN vs. MCI. The architecture is factored into a projection network $P_m:
\mathbb{R}^{6144}\!\to\!\mathbb{R}^{128}$ (the ``biomarker space'') and
a linear classifier $C_m: \mathbb{R}^{128}\!\to\!\mathbb{R}^{2}$. This
factored design is deliberate: by separating the compression step from
the classification step, we obtain a reusable embedding space that the
speech model can later target directly. If the two functions were
entangled in a single monolithic network, there would be no well-defined
intermediate representation for the speech model to align to.
The depth $k$ and widths $\{h_i\}$ are selected by Optuna Bayesian HPO
(60 trials, TPE sampler, MedianPruner). The MRI teacher is trained on 1,228 labeled MRI subjects, providing a substantially larger supervised signal than is available for speech alone. This larger training set is one reason the MRI teacher serves as a reliable reference: its biomarker space is learned from enough examples to generalize well, making it a stable target for the alignment stage.
After training, both $P_m$ and $C_m$ are \emph{frozen}; only these frozen weights are used in Stage~3. Freezing is essential: it ensures that the MRI biomarker space does not shift during alignment training, so that the projection head is forced to move speech embeddings toward MRI rather than the reverse.

\subsection{Stage 3: Cross-Modal Alignment}
The key novelty of MINT is the alignment stage. The frozen speech encoder
(from Stage~1) produces $\mathbf{z}^s$; a trainable projection head
$f_\theta: \mathbb{R}^{128}\!\to\!\mathbb{R}^{96}\!\to\!\mathbb{R}^{128}$
maps it to an aligned embedding $\hat{\mathbf{z}}^s = f_\theta(\mathbf{z}^s)$.
The head uses a single hidden layer with BatchNorm, GELU activation,
Dropout(0.6), and a residual skip connection weighted by 0.1 to ensure
gradient flow. Keeping the projection head deliberately small is a
conscious regularisation choice: with only 74 MCI subjects among the 266 paired training subjects, a larger head would have enough capacity to memorise subject-specific
idiosyncrasies rather than learn a general speech-to-MRI mapping.
The residual connection allows the head to default to a near-identity
mapping early in training, giving the optimiser a stable starting point
before the alignment loss guides it toward the MRI space.

We align normalised embeddings using a two-term loss:
\begin{equation}
  \mathcal{L}_\text{align} =
  \lambda_\text{mse}\,\bigl\|\hat{\mathbf{z}}^s_\text{n} -
    \mathbf{z}^m_\text{n}\bigr\|^2 \;+\;
  \lambda_\text{cos}\,\bigl(1 - \cos\bigl(\hat{\mathbf{z}}^s_\text{n},
    \mathbf{z}^m_\text{n}\bigr)\bigr),
\label{eq:align}
\end{equation}
where subscript $\text{n}$ denotes $\ell_2$-normalisation. The MSE term
penalises magnitude discrepancies and the cosine term penalises directional
misalignment; together they enforce that $\hat{\mathbf{z}}^s$ occupies
precisely the same region of $\mathbb{R}^{128}$ as $\mathbf{z}^m$, so the
frozen linear classifier $C_m$ can be applied to speech without modification.
The core mechanism of MINT relies on the fact that $C_m$, having been trained on $\mathbf{z}^m$, has already established decision boundaries in the MRI biomarker space. By mapping $\hat{\mathbf{z}}^s$ into this space, the speech model can inherit these boundaries instead of rediscovering them from a limited labeled set. Using $\ell_2$-normalised
embeddings for both terms ensures that the loss is not dominated by
subjects with high-magnitude embeddings, which promotes consistent alignment
across the CN and MCI classes. Only $f_\theta$ is updated during Stage~3.
Loss weights $(\lambda_\text{mse}, \lambda_\text{cos})$ and learning rate
are selected by a stratified 5-fold cross-validation grid search on the
training split.

\noindent\textbf{Inference.}
In the \emph{speech-only} scenario, predictions are obtained as
\begin{equation}
 \hat{p} = \mathrm{softmax}(C_m(f_\theta(E_s(\mathbf{x}^s)))),
\label{eq:inference}
\end{equation}
requiring no MRI. In the \emph{paired} scenario, logit-level fusion
averages the logits from $C_m(\mathbf{z}^m)$ and
$C_m(\hat{\mathbf{z}}^s)$, exploiting complementary signal from both
modalities. The two scenarios directly correspond to two different
deployment contexts: a community screening setting where only a smartphone
recording is available, and a memory clinic where MRI has already been
acquired. MINT is designed to be useful in both.
\section{Experiments and Results}
\label{sec:experiments}

\subsection{Dataset and Unified Evaluation Protocol}

\noindent\textbf{Dataset.} We use ADNI-4 Storyteller~\cite{skirrow2024storyteller}: 14,235 unlabeled speech recordings for MAE pretraining, 1,228 MRI-only subjects (677 CN, 551 MCI) for teacher training, and 266 paired subjects (187 CN, 79 MCI) for alignment. Speech features are 209-d acoustic descriptors (pitch, MFCCs, formants, jitter, shimmer, etc.), while MRI features are 6144-d tissue-stratified ResNet-50 embeddings.

\noindent\textbf{Unified evaluation.} All experiments use the same stratified 70/15/15 split $(seed=42)$, yielding 40 test subjects (28 CN, 12 MCI). Methods are evaluated on identical test subjects, and AUC-ROC is used as the primary metric.

\noindent\textbf{Implementation.} All models use AdamW with cosine decay ($T_0=50$), up to 200 epochs, and early stopping based on validation AUC (patience 30). Experiments use a single NVIDIA A100 GPU (40GB).

\subsection{Main Results}

Table~\ref{tab:main} presents a comparison of conventional speech models, including Support Vector Machine (SVM), Multi-Layer Perceptron (MLP), XGBoost, and Logistic Regression (LR), alongside the proposed MINT variants on a unified test set.
\begin{table}[h!]
\centering
\caption{Test-set results. $\dagger$: requires MRI at inference. All methods on identical split.}
\label{tab:main}
\setlength{\tabcolsep}{5pt}
\begin{tabular}{llcc}
\toprule
\textbf{Category} & \textbf{Method} & \textbf{AUC $\uparrow$} \\
\midrule
\multirow{4}{*}{\shortstack[l]{Speech\\baselines}}
  & SVM (RBF) \cite{cortes1995support}                & 0.705  \\
  & MLP from scratch \cite{rumelhart1986learning}     & 0.661  \\
  & XGBoost \cite{chen2016xgboost}                    & 0.613  \\
  & LR on acoustic features \cite{hosmer2000applied}  & 0.580  \\
\midrule
MINT Stage~2 $\dagger$
  & MRI teacher (CV: $0.780\pm0.029$) & 0.958  \\
\midrule
\multirow{2}{*}{MINT (ours)}
  & \textbf{Aligned speech}   & \textbf{0.720}  \\
  & \textbf{Fusion} $\dagger$ & \textbf{0.973}  \\
\bottomrule
\end{tabular}
\end{table}
The speech-only baselines trained on 186 subjects achieve AUCs between 0.580 and 0.705.
SVM achieves the highest AUC (0.705), followed by MLP from scratch (0.661). These results indicate that acoustic features provide discriminative information for early MCI detection, although performance remains modest. The MRI teacher, trained on 1,228 subjects, achieves a 5-fold cross validation AUC of $0.780\pm0.029$ and a test AUC of 0.958. 
The improvement reflects the larger training cohort and the stronger signal from structural neuroimaging.

\begin{figure}[h]
\centering
\setlength{\fboxrule}{0.5pt}
\setlength{\fboxsep}{1pt}

\fbox{\includegraphics[width=0.45\columnwidth]{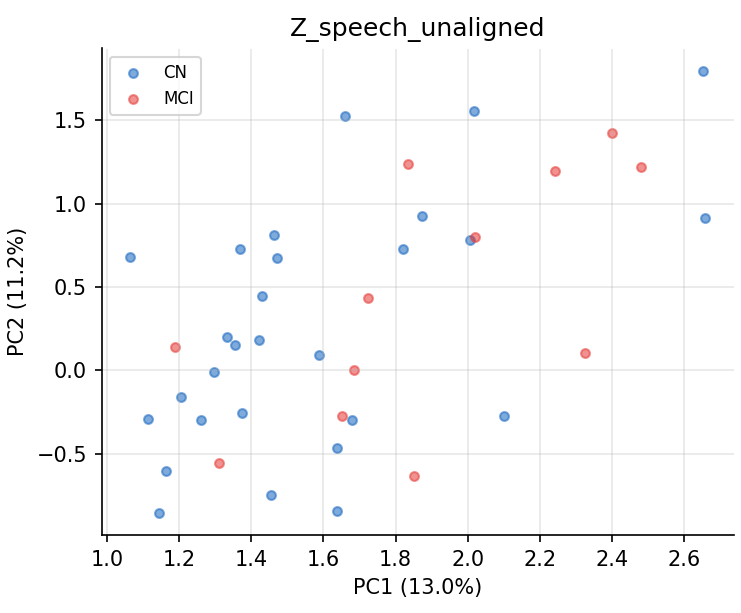}}
\hspace{0.03\columnwidth}
\fbox{\includegraphics[width=0.45\columnwidth]{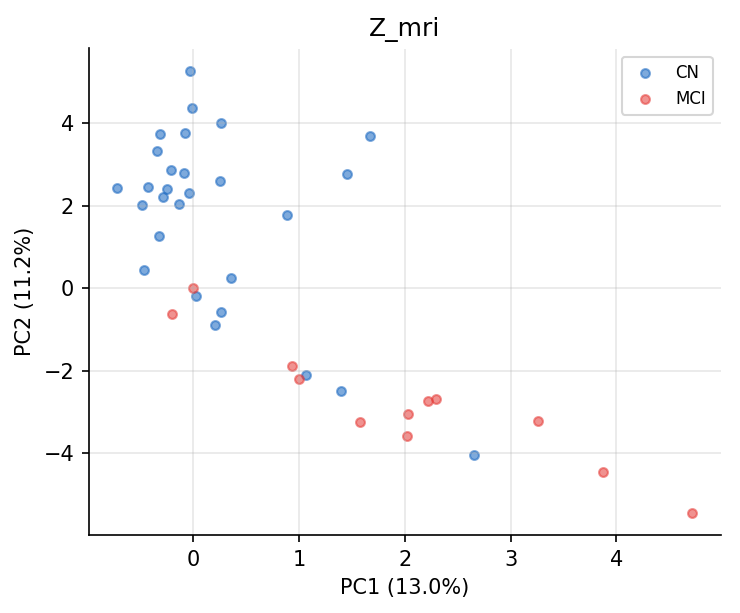}}

\caption{PCA shows greater class overlap in speech embeddings and clearer CN--MCI separation in MRI embeddings.}
\label{fig:alignment}
\end{figure}

\noindent After alignment, MINT speech achieves an AUC of 0.720, compared with 0.705 for the best speech baseline, indicating comparable performance on the held-out test set. During Stage 3, the classifier uses MRI-derived supervision without any additional speech classification loss, indicating that the teacher provides an embedding space for transferring imaging-derived decision structure to speech. Logit-level fusion further improves AUC to 0.973, compared with 0.958 for the MRI teacher, suggesting complementary information from the aligned speech representation. Figure~\ref{fig:alignment} visualizes the modality gap and its alignment using PCA.

\subsection{Ablation Studies}

\noindent\textbf{Self-supervised pretraining.}
Replacing the MAE-pretrained encoder with random initialization reduces speech AUC from 0.720 to 0.667 ($-0.053$) and fusion AUC from 0.973 to 0.907 ($-0.066$), the largest speech-only degradation. This supports MAE pretraining, which learns useful acoustic structure from 14{,}235 unlabeled samples for alignment with only 186 labeled subjects.
\begin{table}[h!]
\centering
\caption{Ablation study on unified test set. All variants use the identical split as Table~\ref{tab:main}.}
\label{tab:ablation}
\setlength{\tabcolsep}{3pt}
\renewcommand{\arraystretch}{1.05}
\begin{tabular}{p{0.38\columnwidth}ccc}
\toprule
\textbf{Variant} & \textbf{Speech} & \textbf{Fusion} & $\boldsymbol{\Delta}$ \\
& \textbf{AUC} & \textbf{AUC} & \textbf{Fusion} \\
\midrule
\textbf{MINT default} & \textbf{0.720} & \textbf{0.973} & --- \\
MSE only ($\lambda_{\mathrm{cos}}=0$) & 0.717 & 0.955 & $-0.018$ \\
Cosine only ($\lambda_{\mathrm{mse}}=0$) & 0.711 & 0.937 & $-0.036$ \\
Larger head (dim 128) & 0.708 & 0.914 & $-0.059$ \\
No pretrain (scratch) & 0.667 & 0.907 & $-0.066$ \\
No dropout (proj. head) & 0.658 & 0.857 & $-0.116$ \\
\bottomrule
\end{tabular}
\end{table}
\noindent\textbf{Dropout regularization.}
Removing Dropout(0.6) causes the largest overall degradation, reducing fusion AUC from 0.973 to 0.857 ($-0.116$). With only 55 MCI-positive training subjects, dropout helps limit subject-specific overfitting.

\noindent\textbf{Loss function design.}
MSE-only and cosine-only losses achieve speech/fusion AUCs of 0.717/0.955 and 0.711/0.937, respectively, below the combined loss (0.720/0.973). This indicates that MSE and cosine capture complementary aspects of embedding alignment.

\noindent\textbf{Projection head capacity.}
Increasing the hidden dimension from 96 to 128 reduces speech and fusion AUC to 0.708 and 0.914, respectively, suggesting overfitting. The 96-d head therefore provides a suitable capacity for the projection.
\label{sec:ablations}

\section{Conclusion}
\label{sec:conclusion}

MINT demonstrates that neuroimaging-derived knowledge can be transferred to speech representations for cognitively grounded MCI detection. The findings indicate that a speech representation aligned to an MRI-derived biomarker space can retain clinically relevant decision structure while being applicable without imaging during inference. This results in two significant conclusions: firstly, large-scale unimodal neuroimaging models can serve as a reliable source of biological supervision for learning from limited paired speech-MRI data; secondly, cross-modal alignment can bridge the gap between accessible speech signals with recognized neuroimaging biomarkers without requiring joint multimodal input during deployment. These results provide preliminary evidence that cross-modal knowledge transfer is a promising direction for speech-based cognitive-impairment screening, while validation on larger and more diverse cohorts remains necessary. Future work will investigate the approach on broader cohorts and temporally informed speech representations to improve robustness and sensitivity to cognitive decline.

\bibliographystyle{IEEEbib}
\bibliography{refs}

@book{world2021global,
  title={Global status report on the public health response to dementia: web annex: methodology for producing global dementia cost estimates},
  author={World Health Organization and others},
  year={2021},
  publisher={World Health Organization}
}

@article{jack2018nia,
  title={NIA-AA research framework: toward a biological definition of Alzheimer's disease},
  author={Jack Jr, Clifford R and others},
  journal={Alzheimer's \& dementia},
  volume={14},
  number={4},
  pages={535--562},
  year={2018},
  publisher={Wiley Online Library}
}

@article{dong2025convolutional,
  title={Convolutional neural network models of structural MRI for discriminating categories of cognitive impairment: a systematic review and meta-analysis},
  author={Dong, Xinxiu and others},
  journal={BMC neurology},
  volume={25},
  number={1},
  pages={400},
  year={2025},
  publisher={Springer}
}

@article{yuan2025magnetic,
  title={Magnetic resonance radiomics-based deep learning model for diagnosis of Alzheimer's disease},
  author={Yuan, Zengbei and others},
  journal={Digital Health},
  volume={11},
  pages={20552076251337183},
  year={2025},
  publisher={SAGE Publications Sage UK: London, England}
}

@inproceedings{thrasher2024te,
  title={TE-SSL: time and event-aware self supervised learning for Alzheimer’s disease progression analysis},
  author={Thrasher, Jacob and others},
  booktitle={MICCAI},
  pages={324--333},
  year={2024},
  organization={Springer}
}

@article{li2024joint,
  title={Joint self-supervised and supervised contrastive learning for multimodal MRI data: Towards predicting abnormal neurodevelopment},
  author={Li, Zhiyuan and others},
  journal={Artificial intelligence in medicine},
  volume={157},
  pages={102993},
  year={2024},
  publisher={Elsevier}
}

@article{fraser2015linguistic,
  title={Linguistic features identify Alzheimer’s disease in narrative speech},
  author={Fraser, Kathleen C and others},
  journal={Journal of Alzheimer’s disease},
  volume={49},
  number={2},
  pages={407--422},
  year={2015},
  publisher={SAGE Publications Sage UK: London, England}
}

@article{chandler2023explainable,
  title={An explainable machine learning model of cognitive decline derived from speech},
  author={Chandler, Chelsea and others},
  journal={Alzheimer's \& Dementia: Diagnosis, Assessment \& Disease Monitoring},
  volume={15},
  number={4},
  pages={e12516},
  year={2023},
  publisher={Wiley Online Library}
}

@inproceedings{zhu2021wavbert,
  title={Wavbert: Exploiting semantic and non-semantic speech using wav2vec and bert for dementia detection},
  author={Zhu, Youxiang and others},
  booktitle={Interspeech},
  volume={2021},
  pages={3790},
  year={2021}
}

@article{li2024useful,
  title={Useful blunders: Can automated speech recognition errors improve downstream dementia classification?},
  author={Li, Changye and others},
  journal={Journal of biomedical informatics},
  volume={150},
  pages={104598},
  year={2024},
  publisher={Elsevier}
}

@article{lin2024multimodal,
  title={Multimodal deep learning for dementia classification using text and audio},
  author={Lin, Kaiying and Washington, Peter Y},
  journal={Scientific Reports},
  volume={14},
  number={1},
  pages={13887},
  year={2024},
  publisher={Nature Publishing Group UK London}
}

@article{hinton2015distilling,
  title={Distilling the knowledge in a neural network},
  author={Hinton, Geoffrey and others},
  journal={arXiv preprint arXiv:1503.02531},
  year={2015}
}

@inproceedings{kunanbayev2024training,
  title={Training vit with limited data for Alzheimer’s disease classification: An empirical study},
  author={Kunanbayev, Kassymzhomart and others},
  booktitle={MICCAI},
  pages={334--343},
  year={2024},
  organization={Springer}
}

@inproceedings{radford2021learning,
  title={Learning transferable visual models from natural language supervision},
  author={Radford, Alec and others},
  booktitle={ICML},
  pages={8748--8763},
  year={2021},
  organization={PmLR}
}

@article{yoon2020vime,
  title={Vime: Extending the success of self-and semi-supervised learning to tabular domain},
  author={Yoon, Jinsung and others},
  journal={NeurIPS},
  volume={33},
  pages={11033--11043},
  year={2020}
}

@inproceedings{devlin2019bert,
  title={Bert: Pre-training of deep bidirectional transformers for language understanding},
  author={Devlin, Jacob and others},
  booktitle={NAACL},
  pages={4171--4186},
  year={2019}
}

@inproceedings{akiba2019optuna,
  title={Optuna: A next-generation hyperparameter optimization framework},
  author={Akiba, Takuya and others},
  booktitle={KDD},
  pages={2623--2631},
  year={2019}
}

@article{zhang2017mixup,
  title={mixup: Beyond empirical risk minimization},
  author={Zhang, Hongyi and others},
  journal={arXiv preprint arXiv:1710.09412},
  year={2017}
}

@article{avants2011reproducible,
  title={A reproducible evaluation of ANTs similarity metric performance in brain image registration},
  author={Avants, Brian B and others},
  journal={Neuroimage},
  volume={54},
  number={3},
  pages={2033--2044},
  year={2011},
  publisher={Elsevier}
}

@article{avants2011open,
  title={An open source multivariate framework for n-tissue segmentation with evaluation on public data},
  author={Avants, Brian B and others},
  journal={Neuroinformatics},
  volume={9},
  number={4},
  pages={381--400},
  year={2011},
  publisher={Springer}
}

@inproceedings{he2016deep,
  title={Deep residual learning for image recognition},
  author={He, Kaiming and others},
  booktitle={CVPR},
  pages={770--778},
  year={2016}
}

@article{tanveer2021classification,
  title={Classification of Alzheimer’s disease using ensemble of deep neural networks trained through transfer learning},
  author={Tanveer, Muhammad and others},
  journal={IEEE Journal of Biomedical and Health Informatics},
  volume={26},
  number={4},
  pages={1453--1463},
  year={2021},
  publisher={IEEE}
}

@article{skirrow2024storyteller,
  title={Storyteller in ADNI4: application of an early Alzheimer's disease screening tool using brief, remote, and speech-based testing},
  author={Skirrow, Caroline and others},
  journal={Alzheimer's \& Dementia},
  volume={20},
  number={10},
  pages={7248--7262},
  year={2024},
  publisher={Wiley Online Library}
}

@article{cortes1995support,
  title={Support-vector networks},
  author={Cortes, Corinna and Vapnik, Vladimir},
  journal={Machine learning},
  volume={20},
  number={3},
  pages={273--297},
  year={1995},
  publisher={Springer}
}

@article{rumelhart1986learning,
  title={Learning representations by back-propagating errors},
  author={Rumelhart, David E and others},
  journal={nature},
  volume={323},
  number={6088},
  pages={533--536},
  year={1986},
  publisher={Nature Publishing Group UK London}
}

@inproceedings{chen2016xgboost,
  title={Xgboost: A scalable tree boosting system},
  author={Chen, Tianqi and Guestrin, Carlos},
  booktitle={KDD},
  pages={785--794},
  year={2016}
}

@book{hosmer2000applied,
  title={Applied logistic regression},
  author={Hosmer, David W and others},
  volume={2},
  year={2000},
  publisher={wiley New York}
}

\end{document}